\documentclass[conference]{IEEEtran}
\IEEEoverridecommandlockouts
\usepackage{cite}
\usepackage{amsmath,amssymb,amsfonts}
\usepackage{algorithmic}
\usepackage{graphicx}
\usepackage{textcomp}
\usepackage{xcolor}
\usepackage{subcaption}
\usepackage{wrapfig}
\usepackage{hyperref}
\usepackage{bookmark}

\def\BibTeX{{\rm B\kern-.05em{\sc i\kern-.025em b}\kern-.08em
T\kern-.1667em\lower.7ex\hbox{E}\kern-.125emX}}

\begin{document}

\title{IMC 2024 Methods \& Solutions Review}

\author{\IEEEauthorblockN{1\textsuperscript{st} Shyam Gupta}
\IEEEauthorblockA{\textit{Technicshe Universitat Dortmund} \\
\textit{Master Student Department of Statistics}\\
Dortmund,Germany\\
shyam.gupta@tu-dortmund.de}
\and
\IEEEauthorblockN{2\textsuperscript{nd} Dhanisha Sharma}
\IEEEauthorblockA{\textit{B.Sc Physics (Honors)} \\
\textit{DAVV (2024)}\\
Indore, India \\
dhanisha522292@gmail.com}
\and
\IEEEauthorblockN{3\textsuperscript{rd} Songling Huang}
\IEEEauthorblockA{\textit{College of Big Data} \\
\textit{Yunnan Agricultural University}\\
Yunnan, China\\
hslingskr@163.com}
}

\maketitle

\begin{abstract}
For the past three years, Kaggle has been hosting the Image Matching Challenge, which focuses on solving a 3D image reconstruction problem using a collection of 2D images. Each year, this competition fosters the development of innovative and effective methodologies by its participants. In this paper, we introduce an advanced ensemble technique that we developed, achieving a score of 0.153449 on the private leaderboard and securing the 160th position out of over 1,000 participants. Additionally, we conduct a comprehensive review of existing methods and techniques employed by top-performing teams in the competition. Our solution, alongside the insights gathered from other leading approaches, contributes to the ongoing advancement in the field of 3D image reconstruction. This research provides valuable knowledge for future participants and researchers aiming to excel in similar image matching and reconstruction challenges.

\end{abstract}


\section{Introduction}
The process of reconstructing 3D models from diverse image collections, known as Structure from Motion (SfM)\cite{sfm}, is critical in Computer Vision but remains challenging, especially with images captured under varied conditions like different viewpoints, lighting, and occlusions. This competition \cite{imc2024} addresses these complexities across six distinct categories:

\begin{enumerate}
  \item \textbf{Phototourism and historical preservation:} Includes diverse viewpoints, sensor variations, and challenges posed by ancient historical sites.
  
  \item \textbf{Night vs. day and temporal changes:} Combines images from different times, lighting conditions, and weather, testing algorithms against temporal variations.
  
  \item \textbf{Aerial and mixed aerial-ground:} Involves images from drones with arbitrary orientations, alongside ground-level shots.
  
  \item \textbf{Repeated structures:} Focuses on disambiguating perspectives of symmetrical objects.
  
  \item \textbf{Natural environments:} Challenges include irregular structures like trees and foliage.
  
  \item \textbf{Transparencies and reflections:} Deals with objects like glassware that lack texture and create reflections, presenting unique computational hurdles.
\end{enumerate}

The competition aims to advance understanding in Computer Vision by bridging traditional image-matching techniques with modern machine learning approaches. By tackling these varied categories, participants contributed to evolving solutions for robust 3D reconstruction from real-world image datasets.

\section{Existing Methods \& Techniques}

Every solution has a unique way of solving the problem. However, there is a standard flow of data most of the solutions followed.
\begin{figure}[htbp]
    \begin{subfigure}[b]{0.45\textwidth}
        \includegraphics[width=\textwidth]{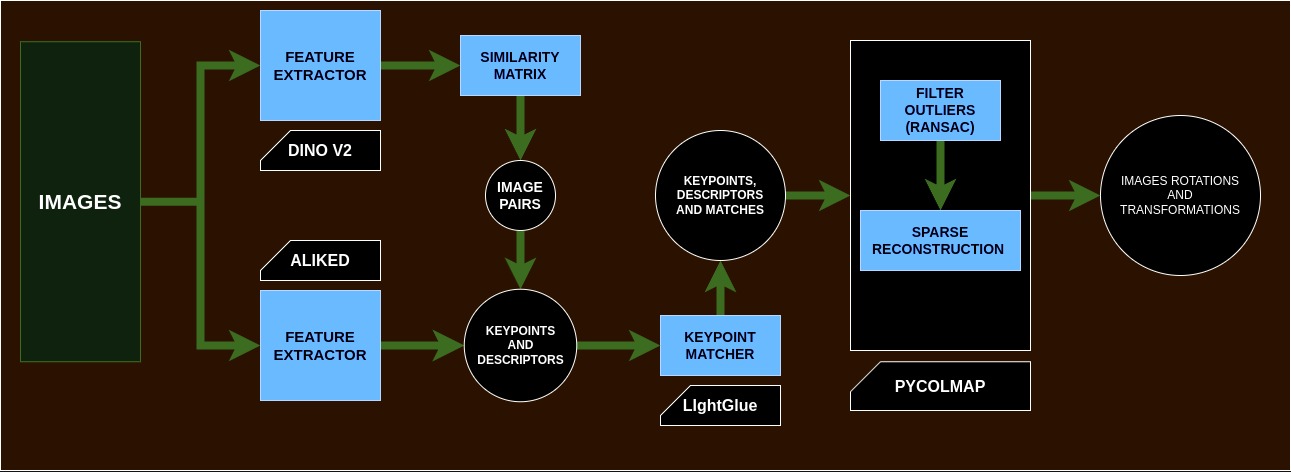}
        \label{fig:using transformations did not help us identify neither helped us to get good results}
    \end{subfigure}
        \caption{General Flow of how data was processed by most of the competitors}

\end{figure}

Following this pipeline, will not result in top ranks itself. Hence, below we mention methods used by top solutions to score higher. In further sections, we discuss how kagglers made a mix'n'match of these techniques to get the best performance.

\section*{MatchFormer \cite{matchformer2022}} 

MatchFormer(2022) was a novel approach to matching multiple views of a scene, crucial for tasks like Structure-from-Motion (SfM), Simultaneous Localization and Mapping (SLAM), relative pose estimation, and visual localization. Traditional methods using detectors and hand-crafted local features are computationally heavy. Recent deep learning methods use Convolutional Neural Networks (CNNs) for feature extraction but are often inefficient due to overburdened decoders.

MatchFormer proposed a new pipeline called extract-and-match, which uses a pure transformer model to perform feature extraction and matching simultaneously. This approach is more intuitive and efficient compared to previous methods. MatchFormer introduces a hierarchical transformer with a matching-aware encoder that uses interleaved self- and cross-attention mechanisms. This design improves computational efficiency and robustness, especially in low-texture scenes.

\section*{DINOv2\cite{dinov2}}

\begin{figure}[htbp]
    \begin{subfigure}[b]{0.45\textwidth}
        \includegraphics[width=\textwidth]{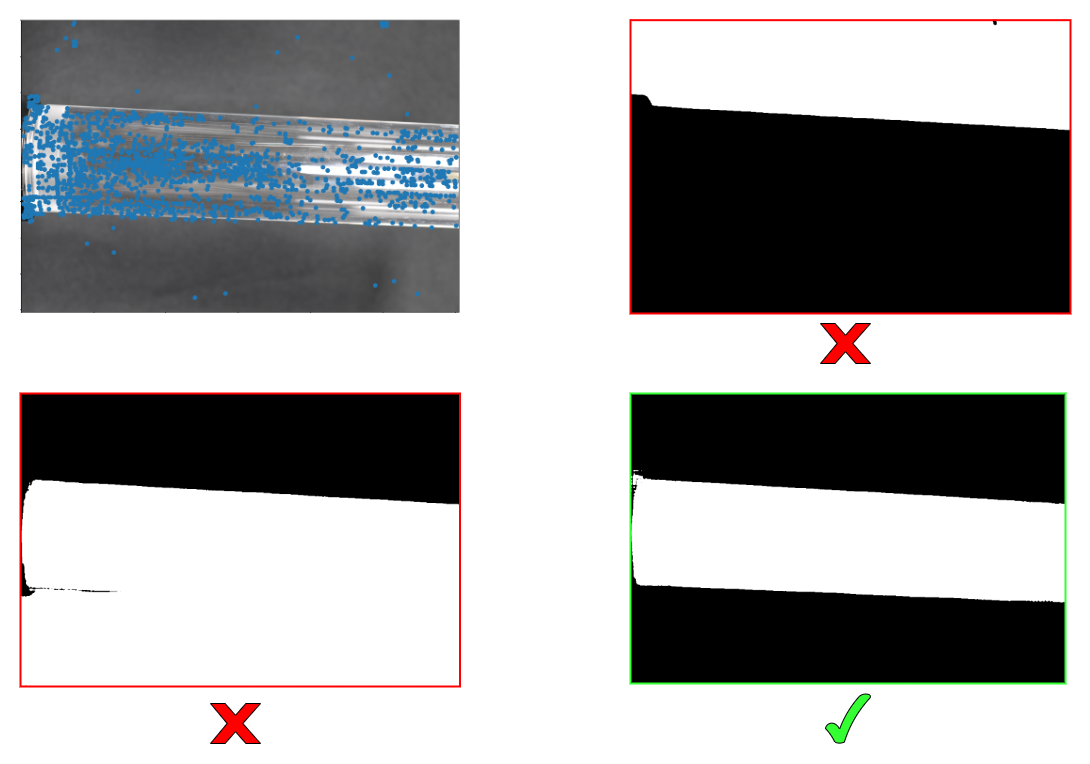}
        \label{using segmentation models significantly improved keypoint detection on transparent}
    \end{subfigure}
        \caption{using segmentation models significantly improved keypoint detection on transparent}

\end{figure}

\begin{figure}[htbp]
    \begin{subfigure}[b]{0.45\textwidth}
        \includegraphics[width=\textwidth]{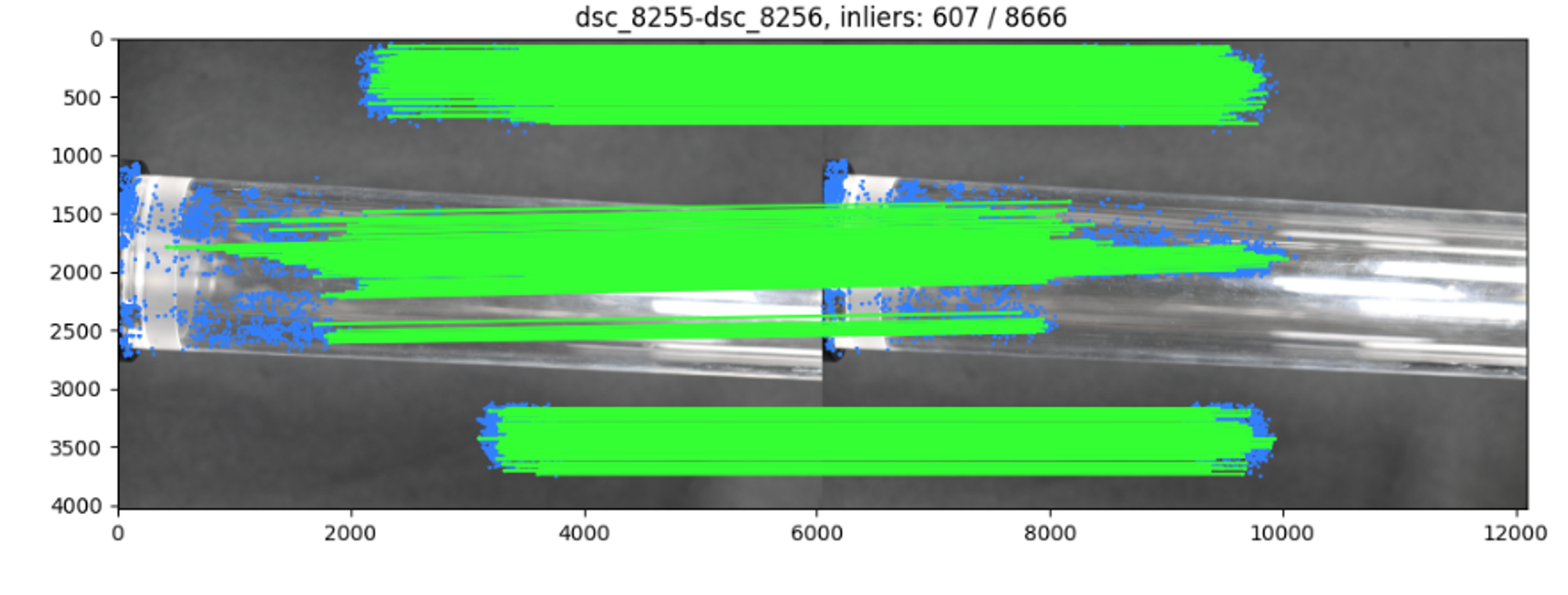}
    \end{subfigure}
        \caption{as clearly visible, background keypoints detected in transparent scene class. To solve this DINOv2 provides seperation between foreground \& background which helps detect foreground keypoints for transparent scene }

\end{figure}

DINOv2 (Distillation of Self-supervised Vision Transformers) enhances segmentation, keypoint detection, and extraction, making it valuable for image matching and 3D reconstruction. DINOv2 leverages self-supervised learning to train vision transformers without labeled data, enabling the model to learn robust and detailed image representations.

For segmentation, DINOv2 uses its learned feature maps to identify and delineate different regions within an image accurately. This segmentation capability is crucial in breaking down complex scenes into manageable parts, aiding in precise object recognition and separation, which is foundational for subsequent processing steps.

In keypoint detection and extraction, DINOv2's ability to generate high-quality feature descriptors ensures that keypoints are distinctive and repeatable. These descriptors are pivotal in matching corresponding points across different images, a core requirement for image matching. The robustness of these keypoints helps in achieving higher accuracy in image alignment and feature matching, which directly impacts the quality of 3D reconstruction.

By integrating DINOv2, image matching algorithms benefit from enhanced feature extraction, leading to more reliable keypoint matches. This improved matching process is essential for constructing accurate 3D models, as it ensures that the spatial relationships between points are preserved across multiple views, resulting in more detailed and accurate 3D reconstructions.

\begin{figure}[htbp]
    \begin{subfigure}[b]{0.45\textwidth}
        \includegraphics[width=\textwidth]{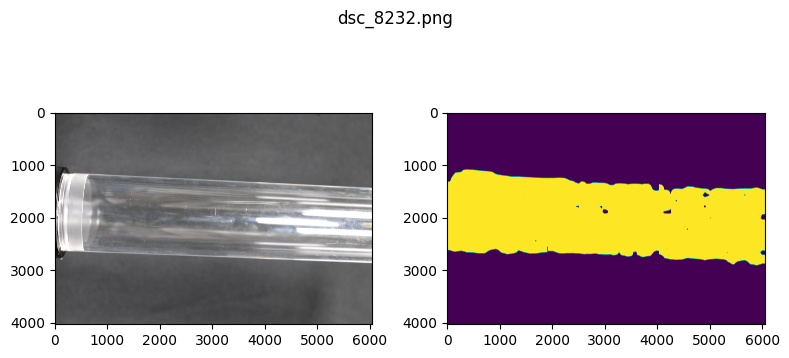}
        \label{figure3}
    \end{subfigure}
        \caption{detection and segmentation of foreground can be seen.}

\end{figure}

\section*{ALIKED\cite{aliked}} 
Efficiently and robustly extracting image keypoints and descriptors is essential for various visual measurement applications, including simultaneous localization and mapping (SLAM)\cite{slam}, computational photography, and visual place recognition. Traditional methods relied on hand-crafted algorithms, which were not very efficient or robust. Modern approaches use deep neural networks (DNNs) for better performance.

\textbf{Keypoints and Descriptors}
\begin{enumerate}
\item
Keypoints: Distinctive points in an image that are used for tasks like image matching and 3D reconstruction.
\item 
Descriptors: Descriptions of the keypoints that allow different keypoints to be compared and matched across images.
\end{enumerate}

Early DNN methods extracted descriptors at predefined keypoints, but newer methods use a single network to extract both keypoints and descriptors simultaneously. These newer methods generate a score map and a descriptor map from which keypoints and descriptors are extracted.

\textbf{Challenges with Existing Methods}
is Existing methods use fixed-size convolutions that lack geometric invariance, which is crucial for accurate image matching. This problem is partially solved by estimating the scale and orientation of descriptors. However, these methods can only handle affine transformations, not more complex geometric transformations. Deformable Convolution Networks (DCNs) can model any geometric transformation by adjusting each pixel's position in the convolution, enhancing descriptor representation. However, DCNs are computationally expensive.

\textbf{Sparse Deformable Descriptor Head (SDDH)} are used to improve efficiency, the paper introduces the Sparse Deformable Descriptor Head (SDDH):

\begin{enumerate}
    \item
        SDDH: Extracts deformable descriptors only at detected keypoints instead of the entire image, significantly reducing computational costs. It uses adjustable positions (offsets) for better flexibility and efficiency in modeling descriptors.
    \item
        ALIKED: A network designed for visual measurement tasks that includes a solution to adapt the neural reprojection error (NRE) loss for sparse descriptors. This adaptation minimizes computational overhead and saves memory during training.
\end{enumerate}

\section*{Dense Matchers and Sparse Keypoint Matchers}

\textbf{Dense Matchers} aim to find correspondences for every pixel or a dense grid of pixels in an image. This comprehensive approach is used in tasks where fine-grained details are important, such as optical flow estimation, depth estimation, and image stitching.

\begin{itemize}
    \item \textbf{Full Coverage}: Consider the entire image, ensuring correspondences for almost every pixel.
    \item \textbf{High Computational Cost}: Require significant computational resources and memory.
    \item \textbf{Applications}: Motion tracking, 3D reconstruction, dense image alignment.
\end{itemize}

\textbf{Sparse Keypoint Matchers}, in contrast, detect and match distinct and repeatable keypoints or features in images. These keypoints are typically corners, edges, or blobs identifiable across different views.

\begin{itemize}
    \item \textbf{Selective Coverage}: Only a subset of points (keypoints) in the image is considered.
    \item \textbf{Lower Computational Cost}: Faster and less computationally demanding than dense matchers.
    \item \textbf{Applications}: Object recognition, image retrieval, feature-based 3D reconstruction.
\end{itemize}

\textbf{Comparison}:
\begin{itemize}
    \item \textbf{Coverage}: Dense matchers cover the entire image, while sparse keypoint matchers focus on specific, informative points. We leverage \underline{Sparse Matchers for this competition} since for SFM we focus on specific object and reject noise.
    \item \textbf{Computational Efficiency}: Sparse keypoint matchers are more efficient computationally. Which proves to be more advantage, since we have 9 hours of runtime limit on Kaggle. 
    \item \textbf{Applications}: Dense matchers are suitable for detailed correspondences, while sparse matchers are ideal for tasks relying on robust and distinctive features.
\end{itemize}

Below we discuss some keypoint matching algorithms, specially \textit{\textbf{LightGlue}}\cite{lightglue} \& \textit{\textbf{OmniGlue}}\cite{omniglue}  since it proved to be most robust and efficient algorithm.

\section*{LightGlue\cite{lightglue}} 

LightGlue is a deep network designed to efficiently and accurately match sparse points between two images. It improves upon SuperGlue by addressing computational limitations while retaining high performance. LightGlue is adaptive, making it faster for easy-to-match image pairs and robust for challenging ones. It is more efficient, easier to train, and suitable for low-latency applications like SLAM\cite{slam} and large-scale mapping.

\begin{figure}[htbp]
    \begin{subfigure}[b]{0.45\textwidth}
        \includegraphics[width=\textwidth]{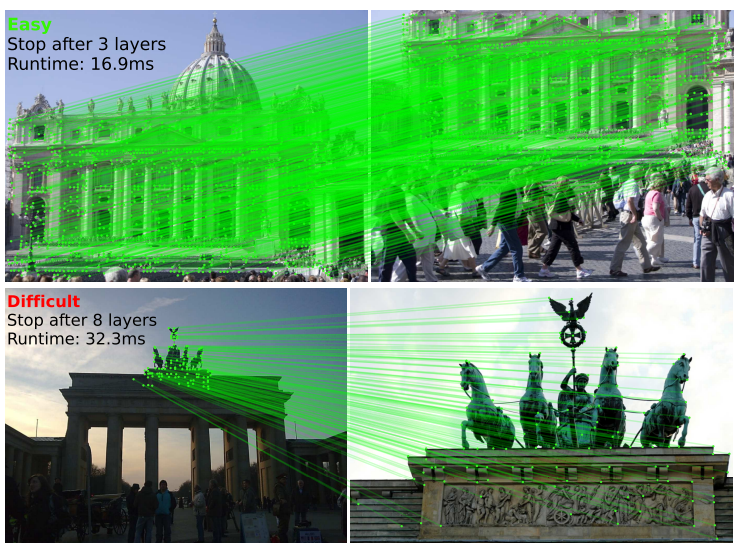}
    \end{subfigure}
        \caption{LigthGlue is faster at matching easy image pairs (top) than difficult ones (bottom) because it can stop at earlier layers when its predictions are confident.}

\end{figure}

\section*{Complex Dense Keypoint Methods}
\subsection{LoFTR (Local Feature TRansformer)}
LoFTR provides dense correspondences between images without needing descriptors. It uses a transformer-based architecture to establish correspondences directly from image patches.
\begin{itemize}
    \item \textbf{Transformer Layers}: Utilizes multi-head self-attention to relate features across the entire image.
    \item \textbf{High Computational Cost}: Despite its accuracy, the dense matching process is computationally intensive.
\end{itemize}

\subsection{SuperGlue\cite{superglue}}
SuperGlue matches keypoints by considering the entire context of both images simultaneously using a graph neural network with attention mechanisms.
\begin{itemize}
    \item \textbf{Graph Neural Network}: Models relationships between keypoints across images.
    \item \textbf{Transformer-Based}: Uses self and cross-attention to enhance matching robustness.
    \item \textbf{Training and Computation}: Requires significant computational resources for training and inference.
\end{itemize}

\subsection*{\textbf{Why LightGlue is Preferred}}
\subsection{Efficiency}
\begin{itemize}
    \item \textbf{Adaptive Matching}: LightGlue adjusts its computational effort based on the difficulty of the image pair, making it faster for easy matches.
    \item \textbf{Early Discarding}: Discards non-matchable points early, reducing unnecessary computations.

While LOFTR (Learning to Optimize Frameworks)\cite{loftr} and Superglue have made significant strides in multimodal research, they do possess certain drawbacks when compared to LightGlue.

\textit{In summary, while LOFTR and Superglue have advanced multimodal research, their drawbacks in terms of complexity, computational requirements, and generalization challenges highlight the potential advantages of LightGlue's approach in certain applications.}

\end{itemize}

\section*{COLMAP \cite{sfm}\cite{colmap}}

\textbf{COLMAP} (Construction and Localization MAPping) is a versatile and widely-used photogrammetry software designed for 3D reconstruction and structure-from-motion (SfM) tasks. It provides a suite of tools for processing images to create 3D models by detecting, describing, and matching keypoints across images, and then using these matches to estimate the 3D structure and camera positions.

\begin{figure}[htbp]
    \begin{subfigure}[b]{0.45\textwidth}
        \includegraphics[width=\textwidth]{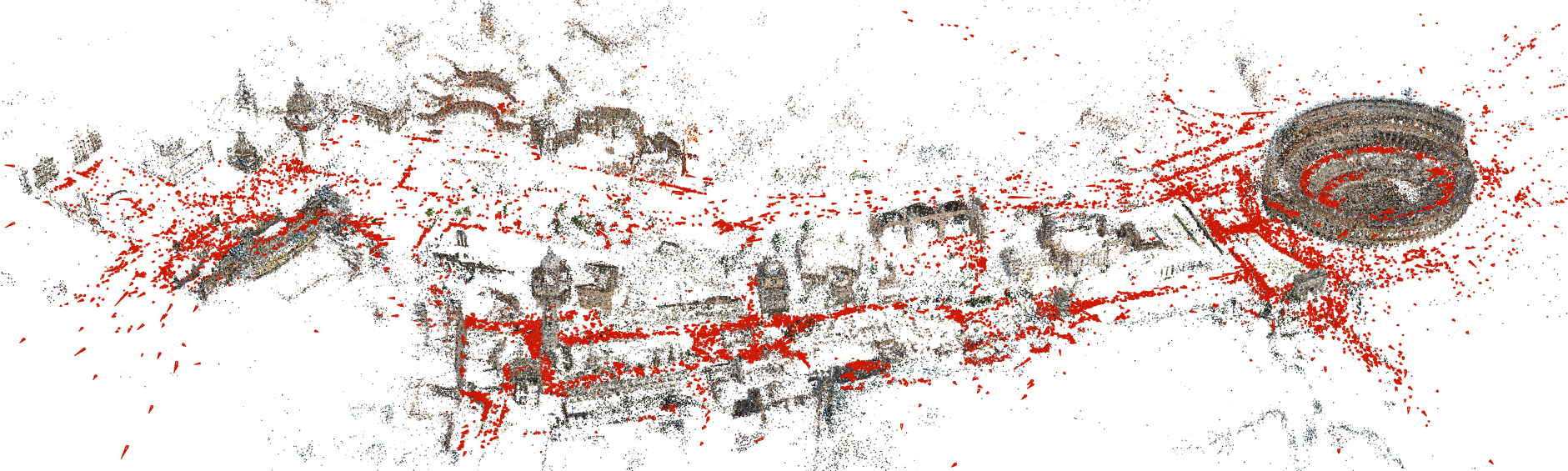}
    \end{subfigure}
        \caption{colmaps reconstruction capabilities- central rome constructed with 21k images.}
\end{figure}

Key features of COLMAP include:
\begin{itemize}
    \item \textbf{Feature Extraction and Matching}: Uses algorithms like SIFT for detecting and matching keypoints across multiple images.
    \item \textbf{Structure-from-Motion (SfM)}: Estimates camera poses and reconstructs sparse 3D points.
    \item \textbf{Multi-View Stereo (MVS)}: Generates dense 3D models by computing depth maps and fusing them into a consistent 3D reconstruction.
    \item \textbf{Scalability}: Efficiently handles large datasets with thousands of images.
    \item \textbf{User Interface}: Provides a graphical interface for easy interaction and visualization, along with command-line tools for automation.
\end{itemize}

COLMAP is favored for its robustness, accuracy, and ease of use, making it suitable for applications in archaeology, architecture, cultural heritage preservation, and more.

\section*{OMNIGLUE\cite{omniglue}}

OmniGlue addresses the generalization limitations of current learnable image matchers, which typically excel in specific domains with abundant training data but falter in diverse, unseen domains. Traditional methods like SIFT, despite being hand-crafted, often outperform these advanced models in unfamiliar contexts due to their domain-agnostic nature.

OmniGlue introduces two key innovations to enhance generalizability:
\begin{enumerate}
    \item \textbf{Foundation Model Guidance}: Utilizes the broad visual knowledge of large pre-trained models like DINOv2\cite{dinov2} to guide the matching process, enhancing performance in domains not covered during training.
    \item \textbf{Keypoint-Position Guided Attention}: Disentangles positional encoding from matching descriptors, avoiding over-reliance on geometric priors from training data, thereby improving cross-domain performance.
\end{enumerate}

Experimental results demonstrate OmniGlue's superior generalization across various domains, including synthetic and real images, scene-level to object-centric, and aerial datasets. Key contributions include:
\begin{itemize}
    \item Enhanced pose estimation accuracy by leveraging foundation model guidance.
    \item Improved cross-domain transferability through innovative positional encoding strategies.
    \item Significant performance gains across diverse datasets, showcasing OmniGlue's robust generalization capabilities.
    \item Ease of adaptation to new domains with minimal fine-tuning data.
\end{itemize}

\subsection*{Comparison with SuperGlue, LightGlue, and LOFTR}

\textbf{SuperGlue\cite{superglue}} is a prominent learnable image matcher that uses attention mechanisms to perform intra- and inter-image keypoint feature propagation, typically leveraging SuperPoint for keypoint detection. While it demonstrates high performance in specific domains, its generalization to unseen domains is limited due to entanglement of local descriptors with positional information, leading to over-specialization.

\textbf{LightGlue\cite{lightglue}} emphasizes lightweight and efficient multimodal fusion, making it suitable for resource-constrained environments or real-time applications. By focusing on simplicity and efficiency, it addresses computational and data requirement issues but may not achieve the same level of performance on diverse datasets as more complex models like OmniGlue.

\textbf{LOFTR\cite{loftr}} (Learning-based Optical Flow with Transformers) employs a coarse-to-fine correspondence prediction paradigm, excelling in dense image matching. However, like other dense matchers, LOFTR struggles with generalization across diverse domains due to its heavy reliance on domain-specific data and computational intensity.

\textbf{OmniGlue\cite{omniglue}}, compared to SuperGlue, LightGlue, and LOFTR, stands out in its generalization capability. By leveraging foundation model guidance and novel keypoint-position attention mechanisms, OmniGlue significantly improves performance in unseen domains while maintaining high accuracy in the training domain. This makes it a more versatile and robust solution for a wide range of image matching tasks, addressing the limitations observed in its predecessors.

\section{Abbreviations and Acronyms}\label{AA}
\begin{enumerate}
    \item \textbf{Correspondences}: Points in one image that match points in another image, allowing the images to be aligned.
    \item \textbf{Sparse Interest Points}: Keypoints in an image that are distinctive and used for matching across images.
    \item \textbf{High-Dimensional Representations}: Numerical descriptions of keypoints that capture their local visual appearance.
    \item \textbf{Robustness}: The ability to handle variations in viewpoint, lighting, and other changes.
    \item \textbf{Uniqueness}: The ability to discriminate between different points to avoid false matches.
    \item \textbf{Transformer Model}: A type of neural network architecture that uses self-attention mechanisms to process input data.
    \item \textbf{Pareto-Optimal}: A state where no criterion (like efficiency or accuracy) can be improved without worsening another.
    \item \textbf{Simultaneous Localization and Mapping (SLAM)}: A technique used in robotics and computer vision to create a map of an environment while simultaneously keeping track of the device’s location within that environment.
    \item \textbf{Self-attention}: A mechanism in neural networks where each element of a sequence pays attention to other elements to understand its context better.
    \item \textbf{Cross-attention}: A mechanism where elements of one sequence pay attention to elements of another sequence, useful in tasks like machine translation and feature matching.
    \item \textbf{Positional Patch Embedding (PosPE)}: A method to incorporate positional information into patches of an image to improve feature detection.
    \item \textbf{Geometric Invariance}: The ability of a method to handle various transformations (like rotation, scaling) in the input data.
    \item \textbf{Deformable Convolution Network (DCN)}: A type of neural network that can adjust the position of each pixel in the convolution, allowing it to model more complex transformations.
    \item \textbf{Neural Reprojection Error (NRE) Loss}: A loss function used to measure the difference between predicted and actual keypoint locations in image matching tasks.
    \item \textbf{Affine Transformations}: Transformations that include scaling, rotation, and translation.
    \item \textbf{Specularities}: Bright spots of light that appear on shiny surfaces when they reflect light sources. These can create difficulties in image matching and 3D reconstruction.
\end{enumerate}

\section{Our Solution}
Our solution aims to implement a complex pipeline of image feature extraction, matching, and 3D reconstruction, integrating a variety of advanced image processing and 3D reconstruction tools. Our solution progressed in 3 steps as follows:
\begin{enumerate}
    \item{}
        The get\_keypoints method uses a deep learning model (such as LoFTR) to extract key points from the image. Then, the matches\_merger method and the keypoints\_merger method are used to merge the key points from different images into a unified dataset to ensure the uniqueness and consistency of the key points. 
    \item{}
        The wrapper\_keypoints method and the reconstruct\_from\_db method use COLMAP to perform 3D reconstruction from the key points and matching data in the database to obtain the camera pose. 
    \item{}
        Finally, the create\_submission method generates a submission file and formats the output results to participate in a specific challenge or competition. The entire pipeline achieves efficient and accurate image processing through accurate feature point extraction, reliable matching filtering, and efficient 3D reconstruction, which is suitable for application scenarios such as autonomous driving, robot navigation, and virtual reality that require sophisticated image processing and 3D reconstruction.
\end{enumerate}

\subsection*{What made the difference?}

Here are a few points we missed on. In Winner's solutions you can observe, they detected these edge cases and solved them efficiently with novel methods, which resulted them better scores.

\begin{enumerate}
    \item{}
        Not correcting image orientation place a significant role. since the algorithms we used are not designed for affine transformations neither they are scale \& rotation invariant.
    \item{}
        We did not solve for transparent \& Low light images. 
    \item{}
        We should have used an ensemble of \textit{aliked+lightglue}\cite{aliked}\cite{lightglue} for key points detection and feature extraction. 

If we would have done these changes. We could have scored higher. 
\end{enumerate}

\section*{\textbf{Top Solutions}}

We have summarized most of the terms and latest research you shold have known for a successful submission in the competition. Top Medal Baggers in Kaggle mostly used permutation \& combination of these techniques to get best scores. 

\section{1st Place Solution}

The final solution combined 3D image reconstruction (I3DR) with \textbf{COLMAP}\cite{colmap} for non-transparent scenes and direct image pose estimation (DIP) for \textit{\textbf{transparent scenes}}. They used an ensemble of \textbf{ALIKED} extractors and \textbf{LightGlue matchers}, cross-validation, multi-GPU acceleration, and a new cropping method. Integration of \textbf{OmniGlue} enhanced match accuracy, and multiple reconstructions were merged for robust results.

\subsection*{Solving Transparent Surface Keypoint Matching}

The solution started with performing orientation correction\cite{orientation}. For detecting keypoints they used ALIKED + LightGlue, However faced with a problem of keypoint detection on transparent surfaces.\textbf{This was a problem which was faced by most of the kagglers. These top solutions discuss and address such problems in detail.}

The initial SfM pipeline with COLMAP \textbf{did not work with transparent scenes}. To address this, they experimented with different strategies, hypothesizing that direct pose estimation might help compute the rotation matrix. They assumed cameras were positioned close to the object, capturing it from all sides.

\subsection*{Approach \#1}
They placed cameras in a circle around the object and sorted images using several methods:
\begin{itemize}
    \item \textbf{Optical Flow}: Calculated magnitude for each image pair and assigned a weight equal to the standard deviation of the magnitude.
    \item \textbf{Pixel-level Difference}: Simple grayscale difference with weights based on the difference value.
    \item \textbf{SSIM Score}: Calculated SSIM index for each pair, assigning a weight of 1 - SSIM.
    \item \textbf{ALIKED+LG Matching}: Number of matches for each pair with weights as 1 / num\_matches.
\end{itemize}
They built a distance matrix from these weights and solved the ordering problem using the Travelling Salesman Problem (TSP).

\subsection*{Approach \#2}
Estimated image order by matching images at high resolution (4096px). The number of matches was higher for consecutive images, using a kNN-like method for estimation.

After this performing reconstruction using COLMAP to get rotation and translation matrix made them rank on TOP of the table, helping them score 0.28, resulting in a gold medal.

\section{2nd Place Solution}
The second-place solution for IMC 2024 devised separate strategies for conventional and transparent scenes due to their distinct characteristics, which were identified through extensive trials.

\section*{Preprocessing}

1. \textbf{Rotation Detection}:
Utilized a rotation detection model to predict and correct image rotations.
Retained original rotations if less than 10\% of images were predicted as rotated, acknowledging the model's potential inaccuracies.\cite{orientation}

2. \textbf{Shared Camera Intrinsics}:
If image dimensions were identical, set all cameras to share the same internal parameters, occasionally improving results by 0.01.

3. \textbf{Transparency Detection}:
Calculated the average difference between images to classify scenes as transparent or not, enabling separate handling for each type.

\section*{Modeling Techniques}

1. \textbf{Global Features}: Developed a robust global feature descriptor combining point and patch features. Extracted point features (\textbf{ALIKED}) and patch features (\textbf{DINO}), establishing one-to-one correspondence based on spatial relationships. Used clustering and the VLAD algorithm to generate global descriptors. This method outperformed existing techniques (NetVLAD, AnyLoc, DINO, SALAD) on VPR-related datasets.

2. \textbf{Local Features}: Utilized three types of local features: \textbf{Dedode v2 + Dual Softmax, DISK + LightGlue, and SIFT + Nearest Neighbor}. The Dedode v2 detector produced rich and evenly distributed feature points. The G-upright descriptor and dual softmax matcher were selected for this purpose.

3. \textbf{MST-Aided Coarse-to-Fine SfM Solution}:
\begin{enumerate}
  \item Constructed a similarity graph with images as vertices and similarities as edges. Computed the Minimum Spanning Tree (MST) to obtain an optimal data association, used for the initial SfM. This stage focused on removing incorrect associations and improving coarse-grained accuracy.
  
  \item Utilized full data associations and the coarse model from Stage 1 to provide initial camera pose priors for geometric verification. This filtered out incorrect feature matches, maintaining coarse-grained advantages while improving fine-grained accuracy.
\end{enumerate}
4. \textbf{Post-Processing}:

Employed pixsfm to optimize the SfM model.
Deployed an HLoc-based relocalization module to process unregistered images, typically resulting in a 0-0.01 score improvement.

\textbf{Handling Transparent Scenes:}Explored various local features (ALIKED, DISK, LoFTR, DKMV3), but none were satisfactory.

Many Kagglers dealt with the problem by separately handling transparent and conventional scenes with tailored preprocessing and modeling techniques, this solution achieved significant accuracy improvements in both types of scenes. Thereby, resulting them in 2nd place gold.

\section{3rd Place Solution}
The third-place solution for IMC 2024, VGGSfM, is a structure-from-motion method based on Visual Geometry Grounded Deep Structure From Motion, which was enhanced for this competition. The approach involved several key strategies:

VGGSfM Across All Frames: Applied VGGSfM to all input frames, improving mAA by 4\% compared to the baseline. However, due to GPU memory limitations on Kaggle servers, this method had to be integrated into the existing pycolmap pipeline.
\begin{figure}[htbp]
    \begin{subfigure}[b]{0.45\textwidth}
        \includegraphics[width=\textwidth]{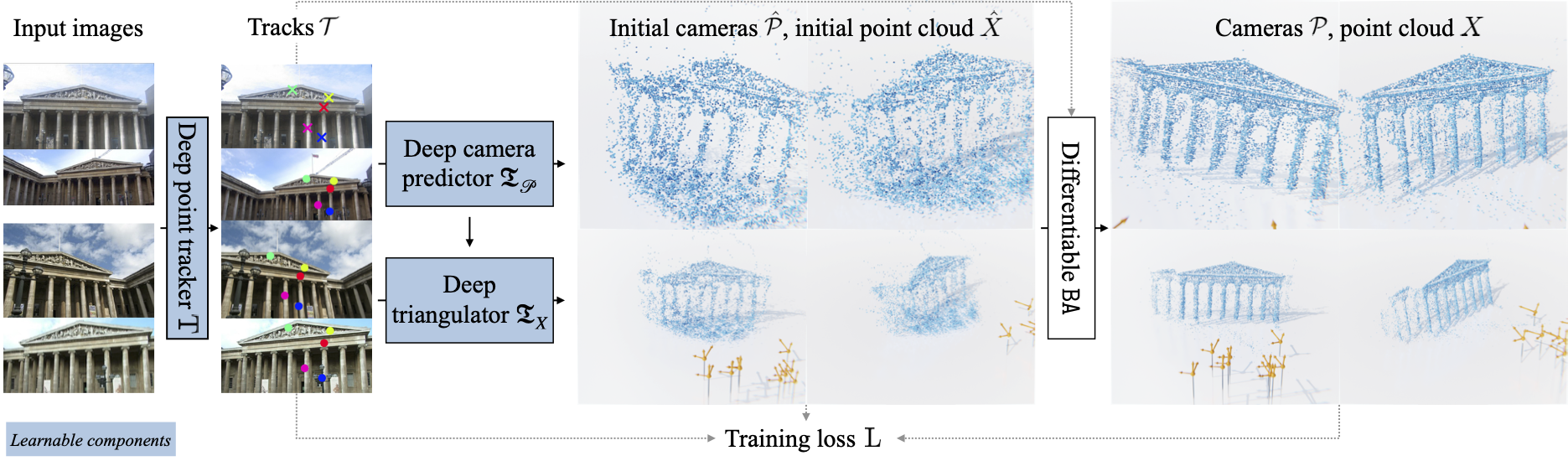}
    \end{subfigure}
        \caption{VGGsfm in action}
\end{figure}

\begin{figure}[htbp]
    \begin{subfigure}[b]{0.45\textwidth}
        \includegraphics[width=\textwidth]{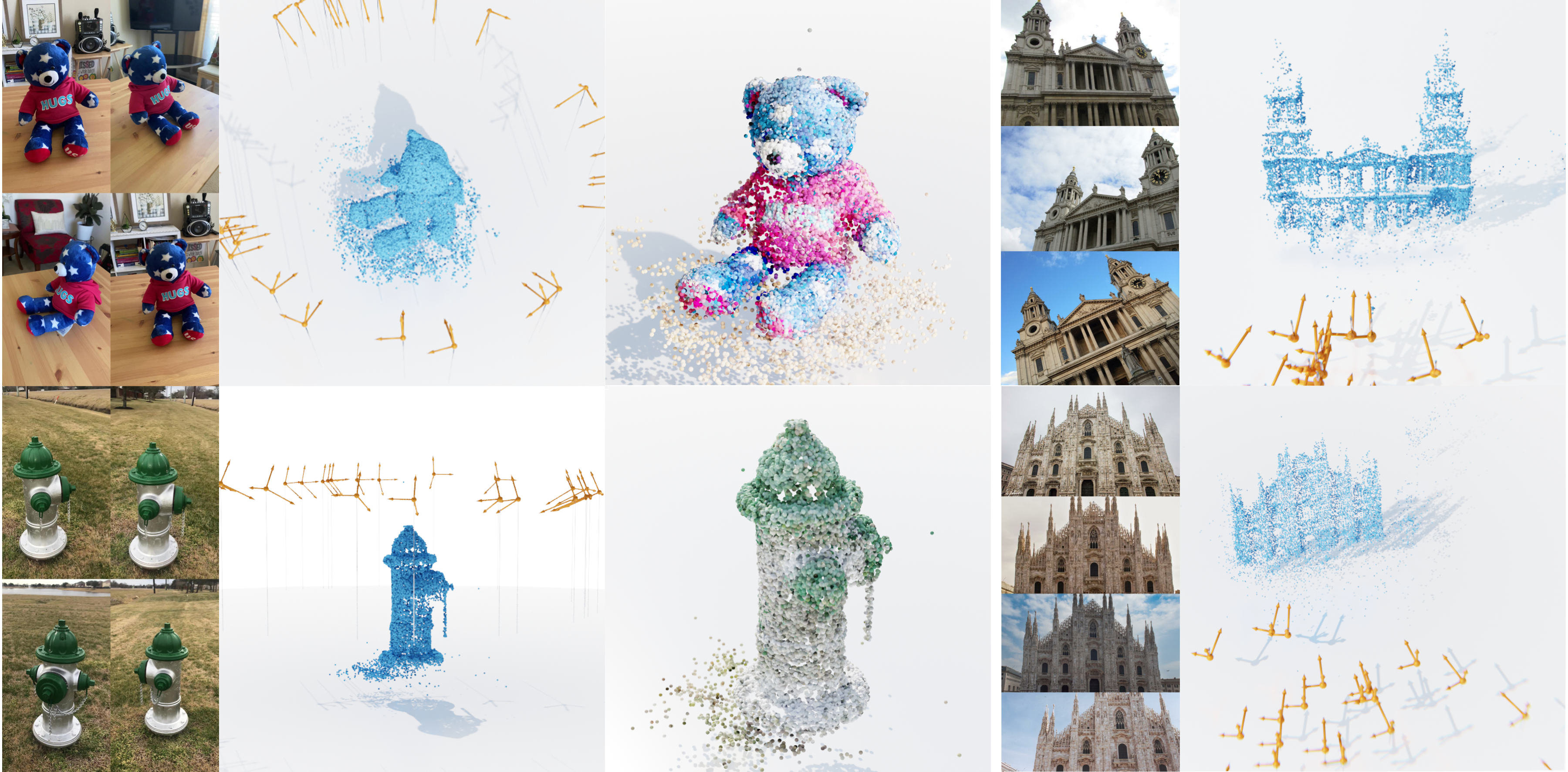}
    \end{subfigure}
        \caption{Reconstruction of In-the-wild Photos with VGGSfM, displaying estimated point clouds (in blue) and cameras (orange).}

\end{figure}
\begin{enumerate}
    \item{}
        Additional Tracks: Utilized VGGSfM's track predictor to estimate 2D matches and fed them into pycolmap. Nearest frames were identified using NetVLAD or DINO V2. This approach improved mAA by 3\% on the evaluation set and public leaderboard score from ~0.17 to ~0.18.
    \item{}
        SfM Track Refinement: Enhanced tracks predicted by pycolmap with VGGSfM's fine track predictor. After running the baseline with ALIKED+LightGlue, 3D points were refined and updated. A global bundle adjustment further optimized camera and point positions, improving mean reprojection error from 0.64 to 0.55 and leaderboard score from ~0.18 to ~0.20.
    \item{}
        Relocating Missing Images: Used VGGSfM to identify and relocate missing images in the scene. This process aligned camera poses and improved the leaderboard score from ~0.20 to ~0.21.
\end{enumerate}
\textbf{Final Solution}: which led to 3rd place.

\begin{enumerate}
    \item{}
        Handled image rotation to maximize matches using a pre-trained model.
    \item{}
        Used matches from both ALIKED+LightGlue and SP+LightGlue.
    \item{}
        For transparent images, extracted an area of interest using DBSCAN on keypoints and ran keypoint detection on these areas again.
\end{enumerate}

\section{4th Place Solution}
\section*{Handling Transparent Images}

\subsection*{Foreground Segmentation}
\begin{itemize}
    \item \textbf{Observation}: Many keypoints in transparent scenes appeared in the background, disrupting camera pose estimation.
    \item \textbf{Solution}: Employed the DINOv2 Segmenter, identifying foreground objects as class5 (``bottle'' in VOC2012). This allowed high-precision segmentation by focusing on transparent objects.
    \item \textbf{Keypoint Detection}: Detected keypoints at the original image scale (1024x1024 grid units) without resizing, which was efficient given the uniform image size. Keypoints were detected only in the foreground area using the segmented results from DINOv2.
\end{itemize}

While looking at accuracte foreground seperation for 4th solution, one should also see this figure ~\ref{figure3} for looking how the method is employed.
\begin{figure}[htbp]
    \begin{subfigure}[b]{0.45\textwidth}
        \includegraphics[width=\textwidth]{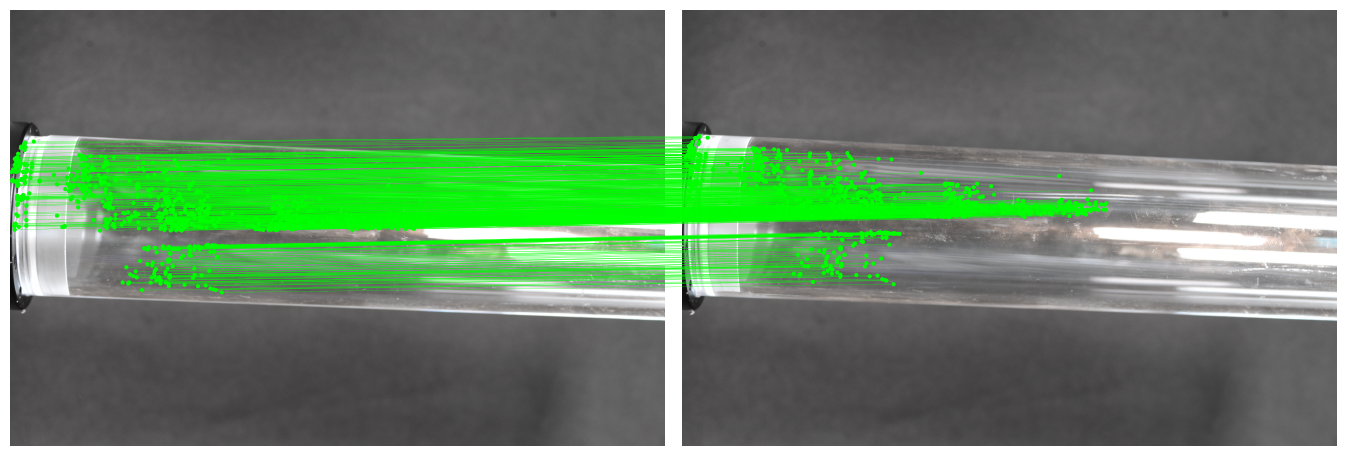}
    \end{subfigure}
        \caption{accurate keypoint detected when using DINOv2}

\end{figure}

\subsection*{Feature Matching}
\begin{itemize}
    \item \textbf{Strategy}: Limited the search for matches to corresponding grids during keypoint detection, significantly reducing the search range and focusing on relevant areas, improving matching efficiency and accuracy.
\end{itemize}

\section*{Leveraging ALIKED and LightGlue}

\subsection*{Non-Transparent Scenes}
\begin{itemize}
    \item \textbf{Keypoint Detection}: Generated keypoints for images rotated in 90-degree increments using ALIKED-n16, retaining keypoints for each rotation.
    \item \textbf{Matching Stage}: Utilized LightGlue to evaluate matches. For each fixed set of keypoints, evaluated matches with rotated counterparts, adopting the combination with the highest number of matches. This ensured robust matching regardless of image rotation.
\end{itemize}

\subsection*{Additional Techniques}

\subsection*{Exhaustive Matching}
\begin{itemize}
    \item Instead of searching for pairs using embedding-based similarity measures like DINOv2 or EfficientNet, exhaustive matching for all image pairs was performed, mitigating the risk of missing matches due to low similarity scores.
\end{itemize}

\subsection*{Using All Images}
\begin{itemize}
    \item By incorporating images beyond those listed in the submission file (up to 100 images for validation), the solution increased the number of triangulated points, enhancing 3D reconstruction accuracy.
\end{itemize}

\section*{Results}
\begin{itemize}
    \item \textbf{Baseline}: Private LB=0.149, Public LB=0.136
    \item \textbf{Add Transparent Trick}: Private LB=0.184, Public LB=0.171
    \item \textbf{Add Exhaustive Matching}: Private LB=0.186, Public LB=0.176
    \item \textbf{Add All Images}: Private LB=0.197, Public LB=0.194
\end{itemize}

These combined approaches led to a robust solution, achieving 4th place in the competition.

\section{5th Place Solution}
\section*{Build Local Evaluation Datasets}
To manage the large dataset realistically, three subsets were created using random sampling strategies. Results across these subsets showed high correlation, validating the chosen subset for local cross-validation (CV) reporting.

\section*{Build a General SfM Pipeline}
The pipeline was divided into three modules:

\subsection{Proposing Pair Candidates by Global Descriptors}
Utilized pretrained models like EVA-CLIP Base, ConvNeXt Base, and Dinov2 ViT Base to extract global features. Customized similarity thresholds based on scene diversity (e.g., Lizard versus Cylinder) using cosine similarity.

\subsection{Matching Pairs in the Candidate List}
Focused on lightweight detector-based methods for image matching due to their efficiency. Experimented with SIFT + NN and LightGlue combinations, noting significant performance boosts in CV but slight drawbacks in LB performance.

\subsection{Reconstruction with Colmap}
Explored various approaches including single-camera usage and manual initial pair settings, aiming for improved reconstruction accuracy. Incremental mapping enhanced consistency in CV results but showed minimal impact on LB.

\section*{Customize the Pipeline for Each Specific Category}

\subsection{Transparent Objects}
Implemented segmentation models like MobileSAM for mask detection and keypoint extraction (e.g., ALIKED). Opted for the smallest mask encompassing most keypoints to focus solely on the object, significantly boosting CV performance with LightGlue.

\subsection{Finding the Best Pairs}
Implemented methods to find consecutive pairs efficiently. Used exhaustive matching for all pairs and built matrices based on match counts, achieving nearly 100\% accuracy in pair selection. But for dark/dim images, I assume that these scenes may have plenty of false positive matches because of their natural properties, so they tried tuning parameters with much more strict values (e.g., increasing the matching threshold to 0.5, etc.). Nonetheless, it didn’t show any improvement. so they tried dopplegangers \cite{doppleganger}. First, they run SfM one time to get all the matching pairs. then used the Doppelganger model to filter out pairs that have high probabilities as false positives (doppelgangers). It didn’t show any improvement on the church scene on the local CV.

\section*{Results}
Leveraging these strategies, the solution consistently improved performance across categories, notably advancing in the LB rankings. Techniques tailored for specific challenges like transparent objects and day-night scenes contributed to the overall success. \textbf{ALIKED+LG+rot+transparentcustom+tuning}:0.195 Which made them rank 5th.

\section{Evaluation Metric- Mean Average Accuracy (mAA)}

Submissions are evaluated based on the mean Average Accuracy (mAA) of the registered camera centers \( C = -R^T T \).

Given the set of cameras of a scene parameterized by their rotation matrices \( R \) and translation vectors \( T \), and the hidden ground truth, the evaluation computes the best similarity transformation \( T \) (scale, rotation, and translation altogether) that registers the highest number of cameras onto the ground truth starting from triplets of corresponding camera centers.

A camera is registered if \( \| C_g - T(C) \| < t \), where \( C_g \) is the ground-truth camera center corresponding to \( C \) and \( t \) is a given threshold. Using a RANSAC-like approach, all possible (N choose 3) feasible similarity transformations \( T' \) derived by Horn's method on triplets of corresponding camera centers \( (C, C_g) \) are exhaustively verified. Here, \( N \) is the number of cameras in the scene.

Each transformation \( T' \) is refined into \( T'' \) by registering the camera centers again using Horn's method, incorporating previously registered cameras with the initial triplets. The best model \( T \), among all \( T'' \) with the highest number of registered cameras, is returned.

\section{Conclusion}
Joining this competition for the first time came with a lot of learning and experiences. We learnt a lot about sfm problems, how edge cases can lead to poor reconstruction results, how using SOTA will not guarantee top score alone, which we saw in case of 2nd place solution.  Special thanks to my co-authors for writing this competition review with me. 

\bibliographystyle{plain}

\vspace{12pt}

\end{document}